\documentclass{aamas2017}
\usepackage{times}
\usepackage{helvet}
\usepackage{courier}
\usepackage{latexsym}
\usepackage{amsfonts,amssymb,amsmath}
\usepackage{alltt}

\usepackage{graphicx}
\usepackage{color}
\usepackage{xspace}

\newcommand{\sat}{\models}

\renewcommand{\phi}{\varphi}
\newcommand{\thm}{\begin{theorem}}
\newcommand{\ethm}{\end{theorem}}
\newcommand{\prf}{\noindent{\bf Proof:} }
\newcommand{\eprf}{\bbox\vspace{0.1in}}
\newcommand{\commentout}[1]{}
\newcommand{\bbox}{\vrule height7pt width4pt depth1pt}
\newcommand{\shortv}{\commentout}
\newcommand{\fullv}[1]{#1}
\newcommand{\F}{{\cal F}}
\newcommand{\K}{{\cal K}}
\renewcommand{\P}{{\cal P}}
\newcommand{\R}{{\cal R}}
\renewcommand{\S}{{\cal S}}
\newcommand{\U}{{\cal U}}
\newcommand{\V}{{\cal V}}
\newtheorem{THEOREM}{Theorem}[section]
\newenvironment{definition}{\begin{DEFINITION} \rm}%
               {\end{DEFINITION}}
\newtheorem{DEFINITION}[THEOREM]{Definition}
\newcommand{\union}{\cup}
\newcommand{\inter}{\cap}
\newcommand{\dfn}{\begin{definition}}
\newcommand{\edfn}{\bbox\end{definition}}
\newenvironment{theorem}{\begin{THEOREM} }%
    {\end{THEOREM}}
\renewcommand{\phi}{\varphi}
\newcommand{\ophi}{\bar{\phi}}

\newcommand{\fpsigma}{$\mbox{FP}^{\Sigma_2^P[\log{n}]}$}
\newcommand{\fpa}{$\mbox{FP}^{{\rm A}[\log{n}]}$}
\newcommand{\fpnp}{$\mbox{FP}^{{\rm NP}[\log{n}]}$}

\newcommand{\en}[1]{\ensuremath{\mathit{en}(#1)}}
\newcommand{\perf}[1]{\ensuremath{\mathit{pf}(#1)}}
\newcommand{\nclob}[1]{\ensuremath{\mathit{nc}(#1)}}
\newcommand{\intd}[2]{\ensuremath{\mathit{in}_{#1}(#2)}}
\newcommand{\dr}{\mathit{dr}}

\newcommand{\postmin}{postcondition minimal\xspace}

\title{Causality, Responsibility and Blame in Team Plans}

\numberofauthors{3}

\author{
\alignauthor
Natasha Alechina\\
\affaddr{University of Nottingham}\\
\email{nza@cs.nott.ac.uk} 
\alignauthor
Joseph Y. Halpern\titlenote{
Supported in part by NSF grants IIS-0534064, IIS-0812045,  
IIS-0911036, and CCF-1214844, and by AFOSR grants 
FA9550-08-1-0438, FA9550-09-1-0266, and FA9550-12-1-0040,
and ARO grant W911NF-09-1-0281.}\\
\affaddr{Cornell University} \\
\email{halpern@cornell.edu}
\alignauthor
Brian Logan\\
\affaddr{University of Nottingham}\\
\email{bsl@cs.nott.ac.uk}
}

\begin{document}
\maketitle

\begin{abstract}
Many objectives can be achieved (or may be achieved more effectively)
only by a group of agents executing a team plan.  
If a team plan fails, it is often of interest to
determine what caused the failure, the degree of responsibility of
each agent for the failure, and the degree of blame attached to each agent.  
We show how team plans can be represented
in terms of structural equations, and then apply the definitions of causality
introduced by Halpern \citeyear{Hal47} and degree of responsibility and 
blame introduced by Chockler and Halpern \citeyear{ChocklerH03} to 
determine the agent(s) who caused the failure and what their degree of
responsibility/blame is.  We also prove new results on the complexity of 
computing causality and degree of responsibility and blame, showing that 
they can be determined in polynomial time for many team plans of interest.
\end{abstract}


%
%



\keywords{Causality; responsibility; blame; team plans}

\section{Introduction} 


Many objectives can be achieved (or may be achieved more effectively) only by 
a coalition or team of agents. In general, for the actions of the agents in the 
team to be successful in achieving the overall goal, their activities must be 
coordinated by a \emph{team plan} that specifies which task(s) should be 
performed by each agent and when they should be performed. 
As with single-agent plans, team plans may fail to achieve their
overall objective: for example, agents may fail to perform a task they have been
assigned. When a failure occurs,
the inter-dependencies between tasks in the team plan can make it
difficult to determine which agent(s) are responsible for the failure:
did the agent simply not perform the task it was assigned, or
was it impossible to perform the task due to earlier failures by other agents?
For example, suppose that a major highway upgrade  does not finish by
the deadline, causing significant traffic problems over a holiday
weekend.
\fullv{Many agents may be involved in the upgrade, 
each executing steps in a large, complex team plan.}
Which agents are the causes of the work not being completed on time?
To what extent are they 
responsible or to blame?

Determining which agents are responsible for the failure of a team
plan is a key step in recovering from the failure, determining which
commitments may have been broken \cite{Singh//:09a} (and hence which sanctions
should be applied), and whether agents should be trusted in the
future \cite{Griffiths//:02a}. 
Identifying those agents most responsible/blameworthy for a plan
failure is useful  for (re)assigning tasks when recovering from the
failure (e.g., we may prefer to exclude agents with a high degree of
blame); if resources are limited, we may wish to focus attention on
the agents most responsible for the failure (e.g., to discover the
reasons for their failure/try to change their behaviour). 
However, there has been relatively little work in
this area. Work in plan diagnosis has focussed on determining the causes of 
failures in team plans (e.g., 
\cite{Micalizio//:04a,Witteveen//:05a}); it typically has not
considered the question of degree of responsibility of agents for the
failure (an exception is the notion of primary and secondary failures
in, e.g., \cite{deJonge//:09a,Micalizio/Torasso:14a}). 
Another strand of work focusses on the problem of how to allocate
responsibility and blame for non-fulfilment of group obligations
(e.g.,  
\cite{Aldewereld//:13a,deLima//:10b,Grossi//:04a,Grossi//:07a,deLima//:10a,Lorini/Schwarzentruber:11a}). 
However, the definitions of causality and responsibility used 
in these work do not always give answers in line with our intuitions
(see, e.g., \cite{Hal48} for examples of what can go wrong).
In this paper, we present an approach to determining the degree of
responsibility and blame of agents for a failure of a team plan based on
the definition of causality introduced
by Halpern \citeyear{Hal47} (which in turn is based on earlier definitions due to Halpern
and Pearl \citeyear{HPearl01a,HP01b}).
\fullv{One advantage of using the Halpern and Pearl  definition of causality}
\shortv{One advantage of using this approach}
is that, as 
shown by Chockler and 
Halpern \citeyear{ChocklerH03}, it can be extended in a natural way to
assign a \emph{degree 
of responsibility} to each agent for the outcome.  
Furthermore, when there is uncertainty about details of what happened,
we can incorporate this uncertainty to talk about
the \emph{degree of blame} of each agent, which is just the expected
degree of responsibility.

We show that each team plan gives rise to a causal model in a natural
way, 
so the definitions of responsibility and blame can be applied without change.  
In addition, it turns out that
the causal models that arise from team plans have a special property:
the equations that characterise each variable are \emph{monotone},
that is,
they can be written as propositional formulas that do not involve negation.
For
such
monotone models, causality for a monotone formula can
be determined in polynomial 
time, while determining the degree of responsibility and blame is NP-complete.
This contrasts with the $D^p$-completeness of 
determining causality
in general \cite{Hal47}
and the $\Sigma^p_2$-completeness of determining 
responsibility (a result proved here).
%
For \emph{\postmin} plans (where preconditions of each step are established
by a unique combination of previous steps), 
the causal models that arise have a further property:
they are \emph{conjunctive}: that is, the equations can be written as
monotone conjunction (so that they have neither negations nor
disjunctions).  In this case, both
causality and degree of responsibility can be determined in polynomial
time.  
\fullv{These complexity results may be
of independent interest.  For example, conjunctive and monotone
formulas are of great interest in databases; indeed, it has already been
shown that for the causal models that arise with databases (which are
even simpler than the conjunctive models that we consider here), computing
causality for conjunctive formulas can be done in polynomial time
\cite{MGMS10a}. (However the notion of causality considered by Meliou at
al.~is closer to the original Halpern-Pearl definition
\cite{HPearl01a}, and thus not quite the same as that considered
here.)}  
\fullv{
This reduction in complexity can be useful in many settings, for
example, where causality, responsibility and blame must be determined
at run-time.} 

The remainder of the paper is structured as follows. In Section
\ref{sec:definitions} we recall the definitions of causality,
responsibility and blame from \cite{ChocklerH03,Hal47}. In Section
\ref{sec:team-plans} we define our notion of team plan, and in Section
\ref{sec:translation} we show how team plans can be translated into
causal models. As noted above, the resulting causal models are monotone;
in Section \ref{sec:complexity} we prove general results on the
complexity of checking causality, degree of responsibility, and degree
of blame for monotone and conjunctive causal models. We conclude
in Section \ref{sec:conclusions}.

\section{Causality, Responsibility, and\\  Blame}\label{sec:definitions}

In this section we briefly review Halpern's definitions of causality
\cite{Hal47} 
and Chockler and Halpern's definition of responsibility and blame
\cite{ChocklerH03}; 
see
\cite{ChocklerH03,Hal47} for further details and intuition.
Much of the description below is taken from \cite{Hal47}.

The Halpern and Pearl approach (hereafter HP) assumes that the world is described in terms of 
variables and their values.  
Some variables may have a causal influence on others. This
influence is modelled by a set of {\em modifiable structural equations}.
It is conceptually useful to split the variables into two
sets: the {\em exogenous\/} variables, whose values are
determined by 
factors outside the model, and the
{\em endogenous\/} variables, whose values are ultimately determined by
the exogenous variables.  The structural equations
describe how the outcome is determined.  

Formally, a \emph{causal model} $M$
is a pair $(\S,\F)$, where $\S$ is a \emph{signature} 
\fullv{that explicitly lists the endogenous and exogenous variables  
and characterises their possible values,} 
and $\F$ is a function that associates a structural equation with each
variable. 
A signature $\S$ is a tuple $(\U,\V,\R)$, where $\U$ is a set of
exogenous variables, $\V$ is a set 
of endogenous variables, and $\R$ associates with every variable $Y \in 
\U \union \V$ a nonempty set $\R(Y)$ of possible values for 
$Y$ (i.e., the set of values over which $Y$ {\em ranges}).  
$\F$ associates with each endogenous variable $X \in \V$ a
function denoted $F_X$ such that $F_X: (\times_{U \in \U} \R(U))
\times (\times_{Y \in \V - \{X\}} \R(Y)) \rightarrow \R(X)$.
%
Thus, $F_X$ defines a structural equation that determines the value
of $X$ given the values of other variables. 
Setting the value of some variable $X$ to $x$ in a causal
model $M = (\S,\F)$ results in a new causal model, denoted $M_{X
\gets x}$, which is identical to $M$, except that the
equation for $X$ in $\F$ is replaced by $X = x$.

Given a signature $\S = (\U,\V,\R)$, a \emph{primitive event} is a
formula of the form $X = x$, for  $X \in \V$ and $x \in \R(X)$.  
A {\em causal formula (over $\S$)\/} is one of the form
$[Y_1 \gets y_1, \ldots, Y_k \gets y_k] \phi$,
where
\fullv{
\begin{itemize}
\item}
$\phi$ is a Boolean
combination of primitive events,
\fullv{\item} $Y_1, \ldots, Y_k$ are distinct variables in $\V$, and
\fullv{\item} $y_i \in \R(Y_i)$.
\fullv{\end{itemize}}
Such a formula is
abbreviated
as $[\vec{Y} \gets \vec{y}]\phi$.
The special
case where $k=0$
is abbreviated as
$\phi$.
Intuitively,
$[Y_1 \gets y_1, \ldots, Y_k \gets y_k] \phi$ says that
$\phi$ would hold if
$Y_i$ were set to $y_i$, for $i = 1,\ldots,k$.

Following \cite{Hal47,HP01b}, we restrict attention here to what are
called {\em acyclic\/} models.  This is the
special case 
where there is some total ordering $\prec$ of the endogenous variables
\fullv{(the ones in $\V$)} 
such that if $X \prec Y$, then $X$ is independent of $Y$, 
that is, $F_X(\vec{z}, y, \vec{v}) = F_X(\vec{z}, y', \vec{v})$ for all $y, y' \in
\R(Y)$.  If $X \prec Y$, then the value of $X$ may affect the value of
$Y$, but the value of $Y$ cannot affect the value of $X$.
If $M$ is an acyclic  causal model,
then given a \emph{context}, that is, a setting $\vec{u}$ for the
exogenous variables in $\U$, there is a unique solution for all the
equations: we simply solve for the variables in the order given by
$\prec$. 

A causal formula $\psi$ is true or false in a causal model, given a
context.
We write $(M,\vec{u}) \sat \psi$  if
the causal formula $\psi$ is true in
causal model $M$ given context $\vec{u}$.
The $\sat$ relation is defined inductively.
$(M,\vec{u}) \sat X = x$ if
the variable $X$ has value $x$
in the
unique (since we are dealing with acyclic models) solution
to
the equations in
 $M$ in context $\vec{u}$ 
\fullv{(i.e., the
unique vector
of values for the exogenous variables that simultaneously satisfies all
equations 
in $M$ 
with the variables in $\U$ set to $\vec{u}$).}
The truth of conjunctions and negations is defined in the standard way.
Finally, 
$(M,\vec{u}) \sat [\vec{Y} \gets \vec{y}]\phi$ if 
$(M_{\vec{Y} = \vec{y}},\vec{u}) \sat \phi$.
Thus, $[\vec{Y} \gets \vec{y}]\phi$ is true in $(M,\vec{u})$ if $\phi$
is true in the model that results after setting the variables in
$\vec{Y}$ to $\vec{y}$.

With this background, we can now give the definition of causality.
Causality, like the notion of truth discussed above,
is relative to a model and a context.
Only conjunctions of primitive events, abbreviated as $\vec{X} =
\vec{x}$, can be causes.  What can be caused are arbitrary Boolean
combinations of primitive events.  Roughly speaking, $\vec{X} =
\vec{x}$ is a cause of $\phi$ if, had $\vec{X} = \vec{x}$ not been the
case, $\phi$ would not have happened.  To deal with many
well-known examples, the actual definition is somewhat more
complicated.

\dfn\label{actcaus}
$\vec{X} = \vec{x}$ is an \emph{actual cause of $\phi$ in
$(M, \vec{u})$} if the following
three conditions hold:
\begin{description}
\item[{\rm AC1.}]\label{ac1} $(M,\vec{u}) \sat (\vec{X} = \vec{x})$ and 
  $(M,\vec{u}) \sat \phi$.
  \item[{\rm AC2$^m$}.]  There is a set $\vec{W}$ of variables in $\V$
and settings $\vec{x}'$ of the variables in $\vec{X}$ and $\vec{w}$
of the variables in $\vec{W}$ such that
$(M,\vec{u}) \sat \vec{W} = \vec{w}$ and
$$(M,\vec{u}) \sat [\vec{X} \gets \vec{x}',
\vec{W} \gets \vec{w}]\neg \phi.$$
\item[{\rm AC3.}] \label{ac3}
$\vec{X}$ is minimal; no subset of $\vec{X}$ satisfies
conditions AC1 and AC2$^m$.
\label{def3.1}  
\end{description}
\end{definition}

AC1 just says that for $\vec{X} = \vec{x}$ to be a cause of $\phi$,
both $\vec{X} = \vec{x}$ and $\phi$ have to be true.  AC3 is a
minimality condition, which ensures that only the conjuncts of
$\vec{X} = \vec{x}$ that are essential are parts of a cause.  AC2$^m$
(the ``m'' is for modified; the notation is taken from \cite{Hal47})
captures the counterfactual.  It says that if we change the value of $\vec{X}$
from $\vec{x}$ to $\vec{x}'$, while possibly holding the values of
the variables in some (possibly empty) set $\vec{W}$ fixed at their
values in the current 
context, then $\phi$ becomes false.  We say that $(\vec{W},\vec{x}')$ is a \emph{witness} to
$\vec{X} = \vec{x}$ being a cause of $\phi$ in $(M,\vec{u})$.
If $\vec{X} = \vec{x}$ is a cause of
$\phi$ in $(M,\vec{u})$ and $X=x$ is a conjunct of $\vec{X} =
\vec{x}$, then $X=x$ is \emph{part of a cause} of $\phi$ in
$(M,\vec{u})$.  

In general, there may be multiple causes for a given outcome.
\commentout{
For example, suppose that we have a vote with three voters.  Let $M$
be a model with endogenous variables $V_1$, $V_2$, $V_3$, and $O$, and
one exogenous variable $U$.  $V_i$
is a binary variable where $V_i = 1$ if voter $i$ votes in
favour, and is 0 otherwise, for $i = 1, 2, 3$;  $O$ 
describes the outcome; $U$ determines how the voters vote.  In a vote
where the outcome is 2--1, so $O=1$, each of the two voters in favour is
separately a cause of $O=1$. 
If the vote is 3--0, then each of $V_1 = 1 \land
V_2 = 1$, $V_1 = 1 \land V_3 = 1$, and $V_2 = 1 \land V_3 =1$ is a
cause of $O=1$.  If $u$ is the context where all three voters
vote in favour, then $(M,u) \sat [V_1 \gets 0, V_2 \gets
  0](O=0)$: flipping the votes of the first two voters will change the
outcome, so AC2$^m$ is satisfied.  Moreover, flipping the vote of just
voter 1 or voter 2 alone does not change the outcome, so AC3 is
satisfied.  Since $(M,\vec{u}) \sat V_1 = 1 \land V_2 = 1 \land O=1$,
AC1 is satisfied as well.  It follows that each of $V_1 = 1$, $V_2 = 1
$, and $V_3 =1 $ is part of a cause of $O=1$ in $(M,u)$.
If we have a vote with 11 voters where the outcome is 6--5, 
then each of the 6 voters who voted
in favour is a cause of the outcome.    If the outcome is 11--0, then
each of the 11 voters who vote in favour is also part of a cause.  
But intuitively, we would like to say that each voter in the latter case is
``less'' of a cause than in the case of the 6--5 victory. 
} 
For example, consider a plan that requires performing two tasks,
$t_1$ and $t_2$. Let $M$ be a model with binary endogenous  variables 
$T_1$, $T_2$, and $Fin$, and one exogenous variable $U$.  
$T_i = 1$ if task $t_i$ is performed 
and 0 otherwise;  $Fin=1$ if the plan
is successfully completed, and 0 otherwise; $U$ determines whether 
the tasks were performed.
\fullv{
(In what follows, we  consider more
sophisticated models where the agents' intentions to perform their 
tasks are determined by $U$.)}
The equation for $Fin$ is
$Fin = T_1 \wedge T_2$.  
If $t_1$ is not performed while $t_2$ is, $T_1 =0$ is the cause of $Fin = 0$.
If $T_1=0$ and $T_2=0$, then both together are
the cause of $Fin= 0$. 
\fullv{Indeed, let $u$ be the context where the two tasks 
are not performed.
AC1 is satisfied since $(M,{u}) \sat T_1=0 \land T_2=0 \land Fin=0$.
AC2$^m$ is satisfied since $(M,u) \sat [T_1\gets 1, T_2 \gets
1](Fin =1)$. Moreover, flipping the value of just
$T_1$ or $T_2$ alone does not change the outcome, so AC3 is
satisfied.} 
If the completion of the plan depended on $n$ tasks instead of two, 
and none of them were performed, the cause would consist of the $n$
non-performed 
tasks.  We would like to say that each of the non-performed tasks was ``less'' 
of a cause of $Fin=0$ than in the case when plan failure is due to a single 
task not being performed.
The notion of \emph{degree of responsibility}, introduced by Chockler
and Halpern \citeyear{ChocklerH03}, is intended to capture this
intuition.  Roughly speaking, the degree of responsibility $X=x$ for
$\phi$ measures the minimal
number  of changes and number of variables that have to be held fixed
in order to make $\phi$ counterfactually depend on 
$X=x$.
We use the formal definition in \cite{Hal48}, which is appropriate for
the modified definition of causality used here.

\dfn
The \emph{degree of responsibility of $X=x$ for $\phi$ in
  $(M,\vec{u})$},
denoted $\dr((M,\vec{u}), (X=x), \phi)$,
is $0$ if $X=x$ is 
not part of a cause of $\phi$ in $(M,\vec{u})$; 
it is $1/k$ if there exists a cause $\vec{X} = \vec{x}$ of $\phi$ 
and a witness $(\vec{W},\vec{x}')$ to $\vec{X} = \vec{x}$
being a cause of $\phi$ in $(M,\vec{u})$ such that 
(a) $X=x$ is a conjunct of $\vec{X} = \vec{x}$,
(b) $|\vec{W}| +
|\vec{X}|= k$, and (c) $k$ is minimal, in that there is no cause $\vec{X}_1
= \vec{x}_1$ for $\phi$ in $(M,\vec{u})$ and witness
$(\vec{W}',\vec{x}_1')$ to $\vec{X}_1 = \vec{x}_1$
being a cause of $\phi$ in $(M,\vec{u})$ 
that includes $X=x$ as a conjunct with $|\vec{W}'| +
|\vec{X}_1| < k$.
\end{definition}

This definition of responsibility 
assumes that everything relevant about
the facts of the world and how the world works is known.  
\fullv{In general, there may be uncertainty both about the context and about
the causal model.}
\shortv{In general, there may be uncertainty about both.}
The notion of \emph{blame} takes 
this into account.
We model an agent's uncertainty by a pair $(\K,\Pr)$, where $\K$
is a set of causal settings, that is, pairs of the form $(M,\vec{u})$,
and $\Pr$ is a probability distribution over $\K$.
We call such a pair an \emph{epistemic state}.
Note that once we have such a distribution, we can talk about the probability that 
$\vec{X} = \vec{x}$ is a cause of $\phi$ relative to $(\K,\Pr)$: it is
just the probability of the set of pairs $(M,\vec{u})$ such that $\vec{X}=\vec{x}$ is a
cause of $\phi$ in $(M,\vec{u})$.  
We also define the \emph{degree of  blame} of $X=x$ 
for $\phi$ to be the expected degree of responsibility:
\dfn The \emph{degree of blame} of $X=x$ for $\phi$ relative to 
the epistemic state
$(\K,\Pr)$ is
$$\sum_{(M,\vec{u}) \in \K}
\dr((M, \vec{u}), X = x, \phi)
\Pr((M,\vec{u})).$$
\end{definition}

\section{Team Plans} \label{sec:team-plans} 

In this section, we define the notion of team plan.
Our definition is essentially the same as 
that used in much of the work in
multiagent planning 
and work in plan diagnosis \cite{Micalizio//:04a,Witteveen//:05a},\fullv{\footnote{In their approach to identifying causes, Witteveen et
  al. \citeyear{Witteveen//:05a} assume that tasks are executed
  as soon as possible, consistent with the order on tasks; we do not
  assume this.} } 
except  
that we explicitly record the assignment of agents to primitive tasks.
It thus encompasses \emph{partial order causal link plans}
\cite{Weld:94a},   
\emph{primitive task networks} \cite{Georgievski/Aiello:15a}, and    
the notion of team plan used in
\cite{Grossi//:04a,Grossi//:07a}, where 
a team plan is constrained to be
a sequence of possibly simultaneous individual actions.

As is standard in planning literature (e.g.,
plans and planning problems relative to a planning domain description; however, for simplicity, we assume that the domain is described using propositional rather than first order logic. 
A \emph{planning domain} is a tuple ${\cal D} = 
(\Pi, {\cal T}$, $\mathit{pre}, \mathit{post})$, where
$\Pi$ is a set of atomic propositions,
${\cal T}$ is the set of tasks possible in the domain, 
and $\mathit{pre}$ and $\mathit{post}$ are functions from ${\cal T}$ to subsets of 
$\Pi \cup \{\neg p: p \in \Pi\}$.
For each $t \in {\cal T}$, $\mathit{pre}(t)$ specifies the
preconditions of $t$ (the set of literals that must hold before $t$
can be executed), and $\mathit{post}(t)$ specifies the postconditions
of $t$ (the effects of executing $t$).

A planning problem ${\cal G}$ is defined relative to a planning domain, and consists of an initial or starting situation and a goal.
The initial situation and goal are specified by the distinguished tasks 
$\mathit{Start}$ and $\mathit{Finish}$ respectively.  
$\mathit{post}(Start)$ is the initial state of the environment, and
$\mathit{Finish}$ has the goal 
as its preconditions and no postconditions.



Given a planning problem, a team plan consists of a set of tasks $T
\subseteq {\mathcal T} \cup \{Start,Finish\}$, 
an assignment of agents to tasks that specifies which agent is going to perform each task in $t \in T \setminus \{\mathit{Start},\mathit{Finish}\}$, and a partial order $\prec$ specifying the order in which tasks in $T$ must be performed.
If $t \prec t'$, whichever agent is assigned to $t$ must get $t$ done before
$t'$ is started.  
$\prec$ is `minimally constraining' in the sense that every
linearization $\prec^*$ of tasks compatible with $\prec$ 
achieves 
the goal (in a sense we make precise below).  
We assume that the agents desire to achieve the goal of the team plan 
and have agreed to the assignment of tasks; we define 
causality and responsibility relative to a team plan.

%
\dfn
A \emph{team plan} $\cal P$ over a planning domain ${\cal D}$
and problem ${\cal G}$
is a tuple ${\cal P} = (T, Ag, \prec$, $\alpha)$, 
where
\begin{itemize}
\item $ \{\mathit{Start},\mathit{Finish}\} \subseteq T \subseteq {\mathcal T} \cup \{Start,Finish\}$ is a finite set of tasks;
\item $Ag$ is a finite set of agents;
\item $\prec$ is an acyclic transitive binary relation on
  $T$ such that $ \mathit{Start} \prec t \prec \mathit{Finish}$ for all
  tasks $t \in T \setminus \{\mathit{Start, Finish}\}$;
\item $\alpha$ is 
  %
     a function that assigns to each task in  $T \setminus
    \{\mathit{Start},\mathit{Finish}\}$ an agent $a \in Ag$ 
    (intuitively, $\alpha(t)$ is the agent 
assigned to execute task $t$; $\mathit{Start}$ 
is executed automatically),
\end{itemize}
such that $\mathit{Finish}$ is executable, that is, the
goal specified by ${\cal G}$ is achieved (in a sense made precise in
Definition~\ref{dfn:accomplish}).
\end{definition}

\commentout{
We assume that the following \emph{completeness} condition  holds for
all team plans; it 
guarantees that the execution of the team plan will achieve the goal
of the plan.  
The completeness condition requires that each task $t \in T$ has an
\emph{establishing set},
where a set $S = \{t_1,\ldots,t_n\}$ of tasks  is an \emph{establishing set for 
task $t$} if and only if
$S$ is a minimal set of tasks that \emph{establishes} all literals $\ell \in
prec(t)$, and a task $t_i$ \emph{establishes literal $\ell$}
if $\ell \in post(t_i)$,
$t_i \prec t$, and there is 
no task $t' \in T$ such that $t_i \prec t' \prec t$ and $\neg \ell \in post(t')$.
} 


Given a task $t$ and a precondition $\ell$ of $t$, a task $t'$ is a
\emph{clobberer}
of $t$ (or the precondition $\ell$ of $t$) if $\sim\!\ell \in post(t')$ 
(where $\sim\!\ell$ denotes $\neg p$ if $\ell = p$ and $p$ if $\ell =
\neg p$).

\dfn\label{dfn:accomplish}
Given a team plan ${\cal P} = (T, Ag, \prec, \alpha)$, 
a task $t' \in T$ \emph{establishes literal $\ell$} for a task $t \in
T$ if $\ell \in prec(t)$, $\ell \in post(t')$, $t' \prec t$, 
and for every task $t'' \in T$ that clobbers $\ell$,
either $t'' \prec t'$ or $t \prec t''$.
A set $S \subseteq T$ of tasks is an
\emph{establishing set  
for task $t \in T$} if and only if
$S$ is a minimal set that establishes all literals $\ell \in
prec(t)$.
${\cal P}$ \emph{achieves the goal specified by ${\cal G}$} if
  each task $t \in T \cup \{\mathit{Finish}\}$ has an establishing set
  in $T$.
\end{definition}




It is easy to check that if ${\cal P}$ achieves the goal and
$\prec^*$ is a linear order on tasks that extends $\prec$ (so that $t
\prec t'$ implies $t \prec^* t'$),
all tasks have their preconditions established at the point when they
are executed.  This justifies the claim that the constraints in
$\prec$ capture all the ordering information on tasks that is needed.

We call a
team plan \emph{\postmin} if there is a unique minimal establishing
set for each task $t \in T$. Most planning
algorithms construct plans that approximate \postmin plans, since they
add only one task for each 
precondition to be achieved. 
However, since they typically do not check for redundancy, the
resulting plan may  
contain several tasks that establish the same precondition $\ell$ of
some task $t$.

As an illustration, consider the plan ${\cal P}_1
= (T_1,Ag_1,\prec, \alpha_1)$, where
$T_1 = \{Start, Finish, t_1,$ $t_2\}$, $t_1$ is laying cables for
traffic signals (under the road surface),
$t_2$ is surfacing the road, $Ag_1 = \{a_1,a_2\}$, $\prec\ =
Start \prec t_1 \prec t_2 \prec Finish$, $\alpha_1(t_1) = a_1$, and
$\alpha_1(t_2) = a_2$. The goal $prec(Finish) = \{c, s\}$, where $c$ stands
for `cables laid' and $s$ for `road surfaced'. $post(Start) = \{\neg c,
\neg s\}$; $prec(t_1) = \{\neg s\}$ (since cables are laid under the surface);
$post(t_1) = \{c\}$; $prec(t_2) = \emptyset$; and $post(t_2) = \{s\}$. 
This plan is accomplishes its goal; the preconditions of $Finish$ are established by
$\{t_1,t_2\}$, while the precondition of $t_1$ is established by $Start$. Note
that $t_2$ is a clobberer of $t_1$
because it undoes the precondition $\neg s$ of $t_1$. For this reason,
$t_2$ is required by $\prec$ to be executed after $t_1$.
%
Note that the plan ${\cal P}_1$ is \postmin.


\commentout{
In what follows, we consider causes of failures of team plans. Some failures are
caused by agents not performing actions they are assigned to perform. For 
generality, we would also like to be able to consider failures that
are caused 
by external events. In order to be able to do this, we generalise the notion of
a team plans by adding events.

\dfn An event-extended team plan $\P = (T,Ag,\prec$, $\alpha,E)$ is a team plan
$\P = (T,Ag,\prec,\alpha)$ together with a possibly empty set of events $E$.
Events have no preconditions, but have postconditions that are sets of
ground literals.
\end{definition}
Events may establish a precondition of a task, or
clobber some task by making a precondition false.
The notion of a teamp plan accomplishing its goal for an
event-extended team plan remains the same:  
each task should be established by other tasks (not by events), and we require 
only that tasks are not clobbered by other tasks.

In what follows, we will refer to event-extended team plans as simply
team plans.
}

\section{Translating Team Plans to \\ Causal Models}\label{sec:translation}

In this section, we apply the definitions of causality, responsibility, and
blame given in Section~\ref{sec:definitions} to the analysis of
team plans.  We start by showing that a team plan 
$\P = (T, Ag, \prec, \alpha)$ determines a causal model $M_{\P}$ in a 
natural way. 
The preconditions of a
task are translated as endogenous variables, as well as whether the
agent intends to perform it.  Whatever determines whether the agent
intends to perform the task is exogenous.  The structural
equations say, for example, that if the agent intends to perform a
task $t$ and all its preconditions 
hold, then the task is performed.

\commentout{
To be able to talk about events in the
environment that affect the execution of the plan, we
\fullv{actually show how to}
construct a causal model
for a team plan $\P = (T, Ag, \prec, \alpha, E)$ extended with a set of 
\emph{events} $E$.
(Of course, $E$ may be empty.)
Events have no preconditions, but have postconditions that are sets of 
ground literals.\footnote{We are not interested in causality between events,
  so we model events as exogenous variables.} 
Events may establish a precondition of a task, or
clobber some task by making a precondition false. 
The of a team plan accomplishing its goal remains the same: each task should be 
established by other tasks (not by events) and we require only that tasks 
are not clobbered by other tasks.
}

For each task $t \in T$, we compute the set $est(t)$ 
and the set  $clob(t)$.  The set $est(t)$
consists of all the establishing sets for task $t$.
The assumption that the plan accomplishes its goal ensures that, for all tasks
$t$, $est(t) \ne \emptyset$.

The set $clob(t)$ contains all pairs $(s,t')$ where $s \in S$ for
some $S \in est(t)$, $s$ establishes some precondition $\ell$ of $t$,
and $t'$ is a clobberer of $\ell$. 

For each task $t \in T$, we have variables
$\en{t}$ for `$t$ is enabled', $\intd{a}{t}$ for `agent $a =
\alpha(t)$ intends to do task $t$', and $\perf{t}$ for `$t$ is performed'.  
$\en{t}$ is true if all the tasks in one of the establishing sets $S$
of $t$ are performed, and no $t'$ such that $(s,t') \in clob(t)$ and
$s \in S$ is performed after $s$ (i.e., $s$ is not clobbered).
(We typically omit $\en{t}$ from the causal model if $est(t)$ is empty, since $\en{t}$ is 
vacuously true in this case.)
In order for $t$ to be performed, it has to be enabled and the
agent assigned the task has to actually decide to perform it; the
latter fact is captured by the formula $\intd{a}{t}$.
For example, even if the roadbed has been laid and
it is possible to surface the road (so the road-surfacing task is enabled), 
if the road-surfacing contractor does not show up, the road will not be 
surfaced.
$\intd{a}{t}$ depends only on the agent $a$.
$\perf{t}$ is true 
if both $\en{t}$ and $\intd{a}{t}$ are true, where $a=\alpha(t)$.
%
%
Finally, for each pair $(s,t')$ in
$clob(t)$, we have a variable $\nclob{s,t',t}$,
which stands for `$t'$ is \emph{not} executed between $s$ 
and $t$'.
Consider again the example plan ${\cal P}_1$ from Section \ref{sec:team-plans}.
The causal model for ${\cal P}_{1}$ has the variables 
$\perf{Start}$, 
$\en{t_1}$, $\intd{a_1}{t_1}$, $\perf{t_1}$, 
$\intd{a_2}{t_2}$, $\perf{t_2}$, $\en{Finish}$, $\perf{Finish}$, 
and $\nclob{Start, t_2, t_1}$.
(Note that we omit $\en{Start}$ and $\en{t_2}$ because $Start$
and $t_2$ have
no preconditions.) 
$\nclob{Start, t_2, t_1}$ is true if $t_2$ is performed
after $t_1$ and false if $t_2$ is performed before $t_1$. $\en{t_1}$
is true if $\perf{Start}$ is true and $\nclob{Start, t_2, t_1}$ is true.

More precisely, a team plan $\P = (T, Ag, \prec, \alpha)$ determines 
causal model $M_\P = ((\U_\P,\V_\P,\R_\P),\F_\P)$
as follows: 
\begin{itemize}
\item $\U_\P \!=\! \{U_{a,t}: t \in T, a = \alpha(t)\} 
\cup \{U_{\nclob{s,t',t}}: s, t', t \in T, (s, t') \in clob(t)\}$.   
Intuitively, $U_{a,t}$ and $U_{\nclob{t',s,t}}$ determine 
the value of $\intd{a}{t}$ and $\nclob{t',s,t}$,
respectively.
\item $\V_\P = 
\{\en{t} : t \in T\} \cup \{\perf{t}: t \in T\} \cup \{\intd{a}{t}: t \in T, a = \alpha(t)\}
\cup \{\nclob{s,t',t}: s, t', t \in T, (s,t') \in clob(t)\}$.
Note that $|\V_{\P}| \le |T|^3 + 3|T|$.
\item $\R_\P(X) = \{0,1\}$ for all variables $X \in \U_\P \union
\fullv{\V_\P$ (i.e.,   all variables are binary).}
\shortv{\V$.}
\item $\F_\P$ is determined by the following equations:\\
$\intd{a}{t} = U_{a,t}$ \\
$\nclob{s,t',t} = U_{\nclob{s,t',t}}$\\
$\perf{t} = \en{t} \wedge \intd{a}{t}$ (where $t \in T$ and $a=\alpha(t)$)\\
$\en{t} = \bigvee\!{}_{S \in est(t)} (\bigwedge\!{}_{s \in S} \perf{s}\!\wedge\!
    \bigwedge\!{}_{(s,t')\in\!clob(t)}\nclob{s,t',t})$. 
\end{itemize}
\vspace{0.5pt}

It should be clear that $M_\P$ captures 
the intent of the team plan $\P$.  In particular,
\fullv{it is easy to see that} the 
appropriate agents performing their tasks results in $\P$
accomplishing its goal iff $(M_\P,\vec{u}) \sat \perf{Finish}$, where
$\vec{u}$ is the context where the corresponding agents intend to
perform their actions
and no clobbering task is performed at the wrong time (i.e., between
the establishing of the precondition they clobber, and the execution
of the task requiring the precondition). 

\fullv{Our causal model abstracts away from pre- and
postconditions of tasks, and concentrates on high level `establishing'
and `clobbering' links between them. This is standard practice in
planning; see, for example, \cite{Weld:94a}.
We also abstract away from the capabilities of agents: our model
implicitly assumes that agents are able to perform the tasks 
assigned to them. All we require is that the preconditions of the task
hold and that the agent intends to perform it. }
\commentout{
In addition, our translation of plans to
causal models relies on two further simplifications. First, we ignore the relationship
between $\perf{t'}$ and $\nclob{s,t',t}$. 
Clearly, $\perf{t'}=0$ entails $\nclob{s,t',t} = 1$ (since if $t'$
is not performed at all, then it is 
not performed after $s$ and before $t$).
Capturing 
the exact relationship between $\perf{t'}$ and $\nclob{s,t',t}$ requires additional variables and does 
change the apportionment of responsibility to agents. 
Second, we assume that 
if $\nclob{s,t',t}=0$ then $en(t)=0$ 
(if some precondition $\ell$ of $t$ established by $s \in S$ is made false by $t'$, 
then $t$ is not enabled). 
It is possible to envisage scenarios
where, for example, $\ell$ could be re-established by another task in $S$ 
or even by some clobberer of $t'$ which has nothing to do with $t$, with the result that  $t$ will
be enabled. 
For example, if $S=\{t_1,t_2\}$, $prec(t)=\{p,q,r\}$, 
$post(t_1)=\{p,q\}$, $post(t_2)=\{p,r\}$, and $\ell=p$, then $t_2$ may re-establish $p$
after $p$ established by $t_1$ is clobbered.
However such an analysis would go beyond the establishing
and clobbering relations that are essential to the plan, and introduce
coincidences that may occasionally prevent plan failure.
Such a fine-grained analysis would involve
modelling each literal in the pre- and postcondition of a task as a
variable. The resulting causal model would be 
much more complex (and non-monotone), but would not change 
the apportionment of responsibility to agents in the case of plan failure.
}

The size of $M_\P$ is polynomial in the size of $\P$ if $\P$ is
postcondition minimal or we treat the maximal number of preconditions
of any task in the plan as a  
fixed parameter
(if there are at most $k$ preconditions of a task, then $est(t)$ has
size at most $2^k$).
Note that all equations are monotone:
there are no negations.  Moreover, 
the only disjunctions in the equations
come from potentially multiple ways of establishing
preconditions of some tasks.
Thus, for \postmin plans 
the formulas are conjunctive.


Having translated team plans to causal models, we can apply the
definitions of Section~\ref{sec:definitions}.  
There may be several causes of $\perf{Finish}$ = $0$.
As we suggested earlier, we are interested only in
causes that involve formulas of the form $\intd{a}{t} = 0$.
We refer to variables of the form $\intd{a}{t}$ as the variables
\emph{controlled} by agent $a$. 

\dfn Agent $a$'s \emph{degree of responsibility for the failure of plan ${\cal P}$}
(i.e., for $\perf{Finish}=0$ in $(M_\P,\vec{u})$, where $M_\P$ is the
causal model determined by a team plan $\P$) is
$0$ if none of the variables controlled by agent $a$ is part of a cause of
$\perf{Finish}=0$ in $(M_\P,\vec{u})$; otherwise,
it is the maximum value $m/k$ such that there exists a cause $\vec{X}=\vec{x}$ 
of $\perf{Finish}=0$ and a witness $(\vec{W},\vec{x}')$ to 
$\vec{X} = \vec{x}$ being a cause of $\perf{Finish}=0$ in $(M_\P,\vec{u})$ 
with $|\vec{X}| + |\vec{W}| = k$, and agent $a$ controls $m$ variables in $\vec{X}$.
\end{definition}

Intuitively, agent $a$'s responsibility is greater if it failed to
perform a greater proportion of tasks. The intentions of agents in our
setting are determined by the context.
Although the intention of some agents can be inferred from observations 
(e.g., if a task $t$ assigned to agent $a$ was performed,
then $\intd{a}{t}$ must hold), in some cases, 
we do not know whether an agent intended to perform a task.
In general, there
will be a set of contexts consistent with the information that we are
given. 
If we are able to define a probability distribution over this set, we can
then determine the degree of blame.   In determining this probability,
we may want to stipulate that, unless we have explicit evidence to the contrary, the
agents always intend to perform their tasks (so that the agents
who we assigned to perform tasks that were not enabled are not to blame). 

To show that our approach gives an intuitive account of
responsibility and blame for plan failures, we briefly outline some
simple scenarios involving the example plan ${\cal P}_1$ and
its corresponding causal model $M_{{\P}_1}$.  
Assume that the context
$u$ is such that $\en{t_1}=1$, $\nclob{Start, t_2, t_1}=1$,
$\perf{t_1}=1$, $\perf{t_2} = 0$, and $\perf{Finish}=0$. We cannot observe the 
values of $\intd{a_1}{t_1}$ and $\intd{a_2}{t_2}$, but from $\perf{t_1}=1$ we 
can conclude that $\intd{a_1}{t_1} = 1$, and, from the fact that $\perf{t_2} =
\intd{a_2}{t_2}$ (since $t_2$ is always enabled), we can conclude
that $\intd{a_2}{t_2} = 0$. Then the cause of $\perf{Finish}=0$
is $\intd{a_2}{t_2} = 0$, and the degree of both responsibility and
blame of agent $a_2$ is 1.
(Note that $\perf{t_2}=0$ is also a cause of $\perf{Finish}=0$, but we
are interested only in causes involving agents' intentions.) 
So far, the analysis is the same as in plan diagnosis: we identify a minimal
set of `faulty components' (unwilling agents) such that, had they
functioned correctly, the failure would not have happened.

For a more complex example of responsibility and blame, consider
a slightly extended plan $\P_2$, which is like $\P_1$, but has an
extra task $t_0 \prec t_1$ that establishes $t_1$: $\en{t_1} = \perf{t_0}$. 
$t_0$ is enabled and assigned to $a_2$. Suppose the context is
$\en{t_0}=1$, $\nclob{t_0, t_2, t_1}=1$,
$\perf{t_0}=0$, $\perf{t_1}=0$, $\perf{t_2} = 0$, and $\perf{Finish}=0$.
As before, $\intd{a_2}{t_0}=0$ and $\intd{a_2}{t_2}=0$ are parts of the cause
of $\perf{Finish}=0$. However, we cannot observe $\intd{a_1}{t_1}$; 
since $t_1$ was not enabled and not performed, we cannot say whether
agent $a_1$ was willing to perform it. In the context $u_1$
where $a_1$ was willing, the cause of $\perf{Finish}=0$
is just $\{\intd{a_2}{t_0}=0,\intd{a_2}{t_2}=0\}$ and the degree of
responsibility of $a_1$ is 0. In the context $u_2$
where $a_{1}$ was not willing, the cause is 
$\{\intd{a_2}{t_0}=0,\intd{a_1}{t_1}=0,\intd{a_2}{t_2}=0\}$ and
$a_1$'s degree of responsibility is 1/3. If we assign
probability 1 to $u_1$, then the blame attached to $a_1$ is 0.


\section{The Complexity of Causality for Monotone Models}\label{sec:complexity}

A causal model is \emph{monotone} if all
the variables are binary and all the equations are monotone (i.e., are
negation-free propositional formulas). A monotone model is
\emph{conjunctive} if all the equations are conjunctive (i.e., they
involve only conjunctions; no negations or disjunctions).
As we have seen, the causal models that are determined by team plans
are monotone; if the team plans are \postmin, then the causal
models are also conjunctive.  

In this section we prove general results on the complexity of checking
causality, degree of responsibility, and degree of blame for monotone
and conjunctive models. 
We first consider the situation for arbitrary formulas.
Recall that the complexity class $D^p$ consists of languages
$L$ such that $L = L_1 \cap L_2$, where $L_1$ is in NP and $L_2$ is
in co-NP \cite{PY}.

\thm
\begin{enumerate}
\item[(a)] \cite{Hal47}
 Determining if $\vec{X} = \vec{1}$ 
  is a cause of $\phi$ 
  in $(M,\vec{u})$ is $D^p$-complete
\item[(b)]
 Determining if $X=x$ is part of a 
  cause of $\phi$ 
  in $(M,\vec{u})$ is $\Sigma^p_2$-complete
\item[(c)] Determining if $X=x$ has degree of responsibility at least
  $1/k$ is $\Sigma^p_2$-complete.
\end{enumerate}
\ethm
\prf Part (a) was proved by Halpern \citeyear{Hal47}.

For part (b), first note that the problem is clearly in $\Sigma_2^p$:
we simply guess $\vec{X}$, $\vec{x}$, $\vec{x}'$, and $\vec{W}$, where
$X=x$ is a conjunct of $\vec{X} = \vec{x}$, compute $\vec{w}$ such that
$(M,\vec{u}) \sat \vec{W} = \vec{w}$ (in general, checking whether
$(M,\vec{u}) \sat \psi$ is easily seen to be in
polynomial time  in acyclic  models, assuming that
the ordering $\prec$ on variables is given, or can be
easily computed from presentation of the equations),
check that $(M,\vec{u}) \sat
[\vec{X} \gets \vec{x}', \vec{W} \gets \vec{w}]\, \neg \phi$, and
 check that there is no
$\vec{Y}\subset \vec{X}$, setting $\vec{y}'$ of the variables in
$\vec{Y}$, and set $\vec{W}'$ such that
$(M,\vec{u}) \sat [\vec{Y}\gets \vec{y}', \vec{W} \gets \vec{w}']\neg \phi$,
where $\vec{w}'$ is such that $(M,\vec{u}) \sat \vec{W}' = \vec{w}'$.

For $\Sigma_2^p$-hardness, we adapt arguments used by Aleksandrowicz et
al.~\citeyear{ACHI14} to show that checking whether a formula satisfies
AC1 and AC2$^m$ 
is $\Sigma_2^p$-complete.
\shortv{Part(c) follows almost immediately from part (b).  Details of
the proofs of parts (b) and (c) can be found 
in the supplementary material. \eprf}

\fullv{
Recall that to show that a language $L$ is $\Sigma_2^p$-hard, it suffices
to show that we can reduce determining if a closed quantified Boolean
formlua (QBF) of the form
$\exists \vec{x}\, \forall \vec{y}\, \phi'$ is true (the fact that it is
closed means that 
all the variables in $\phi'$ are contained in $\vec{x}
\union \vec{y}$)
to checking if a string $\sigma \in L$ \cite{Stock}.    Given a closed QBF
$\phi = \exists \vec{x}\, \forall \vec{y}\, \phi'$, we construct a causal
formula $\psi$, a causal model $M$, and context $\vec{u}$ such that
$\phi$ is true iff $A=0$ is part of a cause of $\psi$ in $(M,\vec{u})$.

 We proceed as follows:  we take $M$ to be a model with endogenous
variables $\V = A \union \vec{X}^0 \union \vec{X}^1 \union
\vec{Y}$, where for each variable $x \in \vec{x}$, there are
corresponding variables $X_x^0 \in \vec{X}^0$ and $X_x^1 \in \vec{X}^1$,
and for each variable $y \in \vec{y}$ there is a corresponding variable
$Y_y \in \vec{Y}$, and a single exogenous
variable $U$.  All the variables are
binary.  The equations are trivial: the value of $U$ determines the values
of all variables in $\V$.
Let $u$ be the context where 
all the variables in $\V$ are set to 0.  
Let $\ophi'$ be the causal formula that results from replacing all
occurrences of $x$ and $y$ 
in $\phi'$ by $X_x^1=1$ and $Y_y=1$, respectively.  Let 
$\psi$ be the formula $\psi_1 \lor (\psi_2 \land \psi_3)$, where
\begin{itemize}
  \item $\psi_1 = \left( \bigvee_{x \in \vec{x}} (X_x^0 = X_x^1)
  \right)$;%
  \footnote{$X_x^0 = X_x^1$ is an abbreviation for the
causal formula $(X_x^0 = 0 \land  X_x^1 = 0) \lor (X_x^0 = 1 \land X_x^1 = 1)$.}
\item $\psi_2 = A=0 \lor \neg(\vec{Y} = \vec{1})$; 
\item $\psi_3 = (A=1) \vee \ophi'$.
\end{itemize}  

We now show that $A=0$ is part of a cause of $\psi$ in $(M,u)$ iff $\phi$ is
true.  First suppose that $\phi$ is true.  Then there is an assignment
$\tau$ to the variables in $\vec{x}$ such that $\forall \vec{y}\, \phi'$
is true given $\tau$.  Let $\vec{x}'$ be the subset of variables in
$\vec{x}$ that are set to true in $\tau$, let $\vec{X}'$ be the
corresponding subset of $\vec{X}^1$ and let $\vec{X}''$ be 
the complementary subset of $\vec{X}^0$ (so that if $x \in \vec{x}$ is false
according to $\tau$, then the corresponding variable $X_x^0$ is in
$\vec{X}''$).
Note that for each variable $x \in \vec{x}$, exactly one
of $X_x^0$ and $X_x^1$ is in $\vec{X}' \union \vec{X}''$.  We claim that  
$A=0 \land \vec{X}'=\vec{0} \land \vec{X}'' = \vec{0} \land \vec{Y} =
  \vec{0}$ is a cause of $\psi$ in 
$(M,u)$.  
  Clearly $(M,u) \sat A=0 \land \psi$
 (since $(M,u) \sat \psi_1$). It is immediate from the definitions
  of $\psi_2$ and $\psi_3$ that
$$(M,u) \sat [\vec{A} \gets 1, \vec{X}' \gets \vec{1}, \vec{X}''
    \gets \vec{1}, \vec{Y}
        \gets \vec{1}] (\neg \psi_1 \land \neg \psi_2),$$ so
$$(M,u) \sat [\vec{A} \gets 1, \vec{X}' \gets \vec{1}, \vec{X}'' \gets
    \vec{1}, \vec{Y}
    \gets \vec{1}] \neg \psi.$$  Thus, AC1 and AC2$^m$ hold.  It suffices to
    prove AC3.  So suppose that there is some subset $\vec{Z}$ of $\vec{A} \union
\vec{X}' \union \vec{Y}$ and a set $\vec{W}$ such that 
$(M,u) \sat [\vec{Z} \gets \vec{1}, \vec{W} = \vec{w}] \neg \psi$,
where $(M,u) \sat \vec{W} = \vec{w}$.  
Since $(M,u) \sat \vec{W} = \vec{0}$, it must be the case that
$\vec{w} = \vec{0}$, so
$(M,u) \sat [\vec{Z} \gets \vec{1}] \neg \psi$.  Clearly we must have
$\vec{Z} \inter (\vec{X}^0 \union \vec{X}^1) = \vec{X}' \union
\vec{X}''$, for otherwise
$(M,u) \sat [\vec{Z} \gets \vec{1}] \psi_1$ and
$(M,u) \sat [\vec{Z} \gets \vec{1}]\psi$.
$(M,\vec{u}) \sat [\vec{Z} \gets \vec{1}]\psi$.
so $(M,u) \sat [\vec{Z} \gets \vec{1}]\psi_3$. 
We must have $A \in \vec{Z}$, since otherwise
$(M,u) \sat [\vec{Z} \gets \vec{1}](A=0)$,
so
$(M,u) \sat [\vec{Z} \gets \vec{1}]\psi_2$, and thus
$(M,\vec{u}) \sat [\vec{Z} \gets \vec{1}]\psi_2$, and thus
$(M,\vec{u}) \sat [\vec{Z} \gets \vec{1}]\psi$.
We also must have $\vec{Y} \subseteq \vec{Z}$, for otherwise
$(M,u) \sat [\vec{Z} \gets \vec{1}]\neg(\vec{Y} = \vec{1})$, and
again $(M,u) \sat [\vec{Z} \gets \vec{1}]\psi_2$ and 
$(M,u) \sat [\vec{Z} \gets \vec{1}]\psi$.
Thus, $\vec{Z} = A \union \vec{X}' \union \vec{X}'' \union \vec{Y}$,
and AC3 holds.

Finally, we must show that
if $A=0$ is part of a cause of $\psi$ in $(M,u)$ then
$\exists\vec{x}\,\forall\vec{y}\,\phi'$ is true.
So suppose that $A=0 \land \vec{Z} = \vec{0}$ is a cause of $\psi$ in
$(M,u)$, where $\vec{Z} \subseteq \V - \{A\}$.
We must have $(M,u) \sat [A \gets 1, \vec{Z} \gets 1]\neg \psi$, which
means that $(M,u) \sat [\vec{Z} \gets 1]\neg \psi_1$.  Thus, for each
$x \in \vec{x}$, $\vec{Z}$ must contain exactly one of $X_x^0$ and
$X_x^1$.  We must also have
$$(M,u) \sat [A \gets 1, \vec{Z} \gets 1](\neg \psi_2
\lor \neg \psi_3).$$  Since $(M,u) \sat [A \gets 1, \vec{Z} \gets 1](A
= 1)$, we have $(M,u) \sat [A \gets 1, \vec{Z} \gets 1]\psi_3$, so
$(M,u) \sat [A \gets 1, \vec{Z} \gets 1]\neg \psi_2$.  It follows that
$\vec{Y} \subseteq \vec{Z}$.  

Let $\nu$ be a truth assignment such that
$\nu(x)$ is true iff $X_x^1 \in \vec{Z}$.  We claim that
$\nu$ satisfies $\forall \vec{y}\, \phi'$.  Once we show this, it
follows that $\phi = \exists \vec{x}\, \forall \vec{y}\, \phi'$ is true,
as desired.
%
Suppose, by way of contradiction, that $\nu$ does not satisfy $\forall
\vec{y}\, \phi'$.    Then there exists a truth assignment $\nu'$ that
agrees with $\nu$ on the assignments to the variables in $\vec{x}$
such that $\nu'$ satisfies $\neg \phi'$.  Let $\vec{Y}'$ be the subset
of $\vec{Y}$ corresponding to the variables $y \in \vec{y}$ that are
true according to $\nu'$.  
Then if
$\vec{Z}'$ is the result of removing from $\vec{Z}$ all the variables
in $\vec{Y}$ that are not in $\vec{Y}'$, we have that
$(M,u) \sat [\vec{Z}' \gets \vec{1}] (\neg \psi_1 \land  \neg \psi_3)$,
so $(M,u) \sat [\vec{Z}' \gets \vec{1}] \neg \psi$.  Thus, 
$A=0 \land \vec{Z} = 0$ is not a cause of $\psi$ (it does not
satisfy AC3), giving us the
desired contradiction.

Part (c) is almost immediate from part (b).  Again, it is easy to see
that checking whether $X=x$ has degree of responsibility in $(M,\vec{u})$ at least
$1/k$ is in $\Sigma^p_2$: we simply guess $\vec{X}$, $\vec{x}$,
$\vec{x}'$, and $\vec{W}$ such that $X=x$ is a conjunct of
$\vec{X} = \vec{x}$ and $|\vec{X}| + |\vec{W}| \le
k$, and confirm that $\vec{X} = \vec{x}$ is a cause of $\phi$ in $(M,\vec{u})$
with witness $(\vec{x}',\vec{W})$.

To show that $X=x$ has degree of responsibility in $(M,\vec{u})$ at least
$1/k$ is $\Sigma_2^p$-hard, given an arbitrary
formula $\phi = \exists \vec{x}\, \forall \vec{y}\, \phi'$. 
 Note that it follows from part (b) that $A=0$ has degree of
responsibility at least $\frac{1}{|\vec{x}| + \vec{y} +1}$ for the
formula $\psi$ as constructed in part (b) iff $\phi$ is true.
The result follows.
\eprf
}


It now follows that by doing binary search
we can compute the degree
of responsibility of $X=x$ for $\phi$ with $\log(|\phi|)$ queries to a
$\Sigma_2^p$ oracle, and, as in \cite{ChocklerH03}, that the
complexity
of computing the degree of responsibility is in \fpsigma, where 
for a complexity class $A$, \fpa\ consists of all 
functions that can be computed 
by a polynomial-time Turing machine with an $A$-oracle 
which on input $x$ asks a total of $O(\log{|x|})$ queries
\cite{Pap84}. 
(Indeed, it is not hard to show that it is \fpsigma-complete; see
\cite{ChocklerH03}.) 
Similarly, the problem of computing the degree of blame is in
$\mbox{FP}^{\Sigma_2^P[n]}$.%
\footnote{We can characterise the complexity of computing the degree of
  blame by allowing parallel 
  (non-adaptive) queries to an oracle (see \cite{ChocklerH03}); we omit this
  discussion here.}

As we now show, checking causality in a monotone model for formulas
$\phi$ or $\neg \phi$, where $\phi$ is monotone, 
is significantly simpler.
For team plans, we are interested in 
determining the causes of $\neg \perf{Finish}$ (why was the plan not completed);
$\perf{Finish}$ is clearly monotone.
Say that a causal model is \emph{trivial} if the equations for the
endogenous variables involve only exogenous variables (so there
are no dependencies between endogenous variables). 

\thm\label{thm:monotone} Suppose that $M$ is a monotone causal model
and $\phi$ is a 
monotone formula. 
\begin{enumerate}
\item[(a)] If $(M,\vec{u}) \sat \phi$, then we can find 
  $\vec{X}$ such that $\vec{X} = \vec{1}$ is a
  cause of $\phi$ in $(M,\vec{u})$ in polynomial time.
\item[(b)] If $(M,\vec{u}) \sat \neg \phi$, then we can find 
  $\vec{X}$ such that $\vec{X} = \vec{0}$ is a
  cause of $\neg \phi$ in $(M,\vec{u})$ in polynomial time.
\item[(c)] Determining if $\vec{X} = \vec{1}$ 
  is a cause of $\phi$ (resp., $\vec{X} = \vec{0}$ 
  is a cause of $\neg \phi$) 
  in $(M,\vec{u})$ can be done in polynomial time.
\item[(d)] Determining if $X=1$ is a part of  a cause of $\phi$ (resp., $X=0$ is
  part of a cause of $\neg \phi$) 
  in $(M,\vec{u})$ is NP-complete;
  NP-hardness holds even if $M$ is a trivial monotone causal model
  and $\phi$ has the form $\psi \land (\phi' \lor X=1)$, where $\phi'$
    is a monotone formula in DNF whose variables are contained
  in $\{X_1, 
  \ldots, X_n, Y_1, \ldots, Y_n\}$ and
  $\psi$ is  the formula   $(X_1 = 1 \lor Y_1 = 1) \land\ldots \land
  (X_n= 1 \lor Y_n = 1)$. 
\item[(e)] Determining if $X=1$ has degree of responsibility at least
  $1/k$ for $\phi$ (resp., $X=0$ has degree of responsibility at least
  $1/k$ for $\neg \phi$)   in $(M,\vec{u})$ is NP-complete.
  NP-hardness holds even if $M$ is a trivial monotone causal model and 
  $\phi$ has the form $\psi \land (\phi' \lor X=1)$, where $\phi'$
  is a formula in DNF whose variables are contained  in $\{X_1,
  \ldots, X_n, Y_1, \ldots, Y_n\}$ and
  $\psi$ is  the formula   $(X_1 = 1 \lor Y_1 = 1) \land\ldots \land
  (X_n= 1 \lor Y_n = 1)$. 
\end{enumerate}
\ethm
\prf
For part (a), let $X_1, \ldots, X_k$ be all the variables that are 1
in $(M,\vec{u})$.  Clearly, only $X_i = 1$ for $i=1, \ldots, k$ can be
part of a cause of $\phi$ in $(M,\vec{u})$ (since $M$ and $\phi$ are
monotone). 
Let $\vec{X}^0 = \{X_1, \ldots,X_k\}$.  Clearly, $(M,\vec{u}) \sat 
[\vec{X}^0 \gets \vec{0}] \neg \phi$. Define $\vec{X}^j$ for $j>0$ inductively
by taking $\vec{X}^j    = \vec{X}^{j-1} - \{X_j\}$ if $(M,\vec{u}) \sat [\vec{X}^j
- \{X_j\}  \gets 0] \neg \phi$, and $\vec{X}^j = \vec{X}^{j-1}$ otherwise. The
construction guarantees that
$(M,\vec{u}) \sat [\vec{X}^k  \gets 0] \neg \phi$, and that
$\vec{X}^k$ is a minimal set with this property.  Thus, $\vec{X}^k =
\vec{1}$ is a cause of $\phi$ in $(M,\vec{u})$.

For part (b), we proceed just as in part (a), except that we switch
the roles of $\phi$ and $\neg \phi$ and replace 0s by 1s.  We leave
details to the reader.

For part (c), to check that $\vec{X} = \vec{1}$ is a cause of $\phi$,
first check
if $(M,\vec{u}) \sat (\vec{X} = \vec{1}) \land \phi$.  (As observed
above, this can be done in polynomial time.)
If so, then AC1 holds.  Then check if
$(M,\vec{u}) \sat [\vec{X} \gets \vec{0}]\neg \phi$.  
If not, $\vec{X} = \vec{1}$ is 
 not a cause of $\phi$ in $(M,\vec{u})$, since AC2$^m$ fails;
the fact that $M$ and $\phi$ are monotone guarantees
that for all sets $\vec{W}$, if $(M,\vec{u}) \sat
\vec{W} = \vec{w}$ and  $(M,\vec{u}) \sat [\vec{X} \gets \vec{0}]\phi$, then 
$(M,\vec{u}) \sat [\vec{X} \gets \vec{0},\vec{W} \gets \vec{w}]\phi$. 
(Proof: Suppose that $W' \in \vec{W}$.  If  $(M,\vec{u}) \sat W' = 1$, 
then, because $M$ and $\phi$ are monotone, $(M,\vec{u}) \sat [\vec{X}
  \gets \vec{0}, W' 
  \gets 1]\phi$.  
On the other hand, if $(M,\vec{u}) \sat W' = 0$, then the fact that
$M$ is monotone
guarantees that $(M,\vec{u}) \sat [\vec{X} \gets \vec{0}](W' = 0)$, so
$(M,\vec{u}) \sat [\vec{X} \gets \vec{0}, W'
  \gets 0]\phi$.%
\footnote{This shows that for monotone causal models and monotone
  formulas, we can always take the set $\vec{W}$ in the witness to be empty.})
For AC3, suppose that $\vec{X} = \{X_1,
\ldots, X_k\}$. Let $\vec{X}_{-i}$ consist of all variables in $\vec{X}$
but $X_i$. Since $M$ and $\phi$ are monotone, it is necessary and
sufficient to show that 
$(M,\vec{u}) \sat [\vec{X}_{-i} \gets \vec{0}]\phi$ for all $i = 1, \ldots,
k$.  Clearly, 
if any of these statements fails to hold, then AC3 does not hold.  On
the other hand, if all these statements hold, then AC3 holds.
This gives us a polynomial-time algorithm for checking if $\vec{X} =
\vec{1}$ is a cause of $\phi$ in $(M, \vec{u})$.  The algorithm for
checking that $\vec{X} = \vec{0}$ is a cause of $\neg \phi$ is
essentially the same, again replacing $\phi$ by $\neg \phi$ and
switching the role of 0 and 1.

For part (d), checking if $X=1$ is part of a cause of $\phi$ in
$(M,\vec{u})$ is clearly in NP: guess a cause $\vec{X} = \vec{1}$
that includes $X=1$ as a conjunct, and confirm that it is a cause as
discussed above.  

To show that checking if $X=1$ is part of a cause of $\phi$ in
$(M,\vec{u})$ is NP-hard, suppose that we are given a propositional
formula $\phi$,  
with primitive propositions 
$x_1, \ldots, x_n$.  Let $\phi^r$ be the result of (i) converting $\phi$
to \emph{negation normal form}\fullv{ (so that all the negations are driven
in so that they appear only in front of primitive propositions---this
conversion can clearly be done in polynomial time, indeed, in linear
time if $\phi$ is represented by a parse tree)} and (ii) replacing 
all occurrences of $\neg x_i$ by $y_i$, where $y_i$ is a fresh
primitive proposition.
Note that $\phi^r$ is monotone.   
(The formula $\phi^r$ was first introduced by Goldsmith, Hagen, and Mundhenk
\citeyear{Goldsmith//:08a} for a somewhat different purpose.)
\commentout{
The following two facts are easy to check:
  \begin{itemize}
    \item $\phi$ is satisfiable iff there is a satisfying assignment
      for $\phi^c$ where exactly one of $x_i$ and $y_i$ is true, for
      $i = 1, \ldots, n$;
      \item if $\phi$ is unsatisfiable, then $\phi$ is satisfiable only
        by an assignment where at least one of $x_i$ and $y_i$ is
        true, for all $i \in\{ 1, \ldots, n\}$, and both $x_j$ and $y_j$ are
        true for some $j \in \{1, \ldots, n\}$.
  \end{itemize}
}

Let $\ophi^r$
be the monotone causal formula that results by
replacing each occurrence of $x_i$ (resp., $y_i$) in $\phi^r$
by $X_i = 1$ (resp., $Y_i = 1$).  
Let $\ophi^+ = \psi \land (\ophi^r \lor X=1)$, where $\psi$ is
$$(X_1 = 1 \lor Y_1 = 1) \land\ldots \land (X_n= 1 \lor Y_n = 1).$$
    Let $M$ be a model where $\V = \{X, X_1, \ldots, X_n, Y_1,
    \ldots, Y_n\}$ and $U$ is the only exogenous
    variable.  $U$ determines the values of all the variables in $\V$,
    so again there are no interesting equations.  
    Let $u$ be the context
    where all these variables are 1.  
    We claim that $X=1$ is part of a
    cause of $\ophi^+$ in $(M,u)$ iff $\neg \phi$ is satisfiable.
        This clearly suffices to prove the NP lower bound
        (since $\phi$ is satisfiable iff $X=1$ is a cause of
        $\bar{\neg \phi}^r$ in $(M,u)$).
 \shortv{We omit details due to lack of space.}
    To prove the claim, first suppose that $\neg \phi$ is
    unsatisfiable, so $\phi$ is valid.
    Let $\vec{Z}$ be a subset of $\{X_1,\ldots, X_n, Y_1, \ldots, Y_n\}$.
We claim that if $\vec{Z}$ contains at most one of $X_i$ and $Y_i$ for
$i=1, \ldots n$, then $(M,u) \sat
[\vec{Z} \gets \vec{0}](\psi \land \ophi^r)$.  
The fact that $(M,u) \sat [\vec{Z} \gets \vec{0}]\psi$ is immediate.
To see that $(M,u) \sat [\vec{Z} \gets \vec{0}]\ophi^r$, first suppose
that $\vec{Z}$ contains exactly one of $X_i$ or $Y_i$ for all $i \in
\{1,\ldots, n\}$. Then $\vec{Z}$ determines a truth assignment to $\vec{x}$ in
the obvious way, so  $(M,u) \sat [\vec{Z} \gets \vec{0}]\ophi^r$,
since $\phi$ is valid.
Since $\ophi^r$ is monotonic, it follows that if $\vec{Z}$ contains at
most one of $X_i$ or $Y_i$ for all $i \in \{1, \ldots, n\}$, then we
must also have 
$(M,u) \sat [\vec{Z} \gets \vec{0}]\ophi^r$.  This completes the argument.

Now suppose, by way of contradiction, that $X=1$ is part of a cause of
$\ophi^+$ in $(M,u)$.   Then there exists a subset
$\vec{Z}$ of $\{X_1, \ldots, X_n$, $Y_1, \ldots, Y_n\}$ such
that $(M,u) \sat [\vec{Z} \gets \vec{0}, X \gets 0]\neg \ophi^+$.
By the argument 
above, it cannot be the case $\vec{Z}$ contains at most one of
$X_i$ and $Y_i$ for all $i = 1, \ldots, n$, for otherwise, we
must have
$(M,u) \sat [\vec{Z} \gets \vec{0}, X \gets 0](\psi \land \ophi^r)$,
and hence $(M,u) \sat \mbox{$[\vec{Z} \gets \vec{0}, X \gets 0]\ophi^+$}$.
Thus, it must be the case that $\vec{Z}$ includes both $X_i$ and $Y_i$
for some $i \in \{1,\ldots, n\}$.  But then
$(M,u) \sat \mbox{$[\vec{Z} \gets \vec{0}]\neg \psi$}$, so
$(M,u) \sat [\vec{Z} \gets \vec{0}]\neg \ophi^+$, which contradicts AC3.
Thus, $X=1$ is not part of a cause of $\ophi^r$ in $(M,u)$.

Now suppose that $\neg \phi$ is satisfiable.
Then there is a set $\vec{Z} \subseteq \{X_1,$ $\ldots$, $X_n, Y_1,
\ldots, Y\}$ that includes 
exactly one of $X_i$ and $Y_i$, for $i=1, \ldots, n$, such that
$(M,u) \sat [\vec{Z} \gets \vec{0}]\neg \ophi^r$.  Let $\vec{Z}'$ be a
minimal subset of $\vec{Z}$ such that 
$(M,u) \sat [\vec{Z}' \gets \vec{0}]\neg \ophi^r$.  We claim that
$\vec{Z}' = 1 \land X=1$ is a cause of $\ophi^+$.
AC1 trivially holds.
Clearly
$(M,u) \sat [\vec{Z}' \gets \vec{0}, X \gets 0]\neg (\ophi^r \lor
X=1)$, so $(M,u) \sat \mbox{$[\vec{Z}' \gets \vec{0},  X \gets 0]\neg
\ophi^+$}$ and AC2 holds.  By choice of $\vec{Z}'$, there is no strict subset
$\vec{Z}''$ of $\vec{Z}$ such that
$(M,u) \sat [\vec{Z}'' \gets \vec{0}]\neg \ophi^r$.  Since $\vec{Z}'$
contains at most one of $X_i$ or $Y_i$ for $i = 1, \ldots, n$, we have
that $(M,u) \sat [\vec{Z}' \gets \vec{0}]\psi$.  It now easily follows
that AC3 holds.  Thus, $X=1$ is part of a cause of $\ophi^+$.

Since to get NP-hardness it suffices to consider only CNF formulas, and
the result above shows that $X=1$ is a cause of $\phi^+$ iff $\neg \phi$ is
satisfiable, we can restrict to $\phi$ being a DNF formula.  The model
$M$ is clearly a trivial monotone model.  This completes the proof of
part (d).

The argument that determining if $X=0$ is a part of  a cause of $\neg
\phi$ is NP-complete is almost identical.  In particular, 
essentially the same argument as that above shows that
$\neg \phi$ is a satisfiable propositional formula iff $X=0$ is part
of a cause of 
$\neg \ophi^+$ in $(M,u')$, where $M$ is as above and $u'$ is the
context where all variables in $\V$ get value 0.

Part (e) follows easily from part (d).
\shortv{We omit details. \eprf}
\fullv{To show that checking if
the degree of responsibility of $X=1$ for $\phi$ is at least
$1/k$ is in NP, given $k$,  
we guess a cause $\vec{X} = \vec{1}$ that includes $X=1$ as a conjunct
and has $k$ or fewer conjuncts.  As observed above, the fact that
$\vec{X} = \vec{1}$ is a cause of $\phi$ in $(M,\vec{u})$ can be
confirmed in polynomial time.

For the lower bound, using the notation of part (d), if the
propositional formula $\phi$ mentions $n$ primitive propositions, say
$x_1, \ldots, x_n$, then we claim that $X=1$ has degree of
responsibility at least $1/(n+1)$ for $\ophi^+$
in $(M,u)$ iff $\neg \phi$ is satisfiable. 
As observed above, if $\neg \phi$ is not satisfiable, then $X=1$ is not a
cause of $\neg \ophi^+$, and hence has degree of responsibility 0.  On
the other hand, if $\neg \phi$ is satisfiable, then as shown above, $X=1$
is part of a cause $\vec{Z}^+ = 1$ for $\phi$ in $(M,u)$.
Since $|\vec{Z}^+| = n+1$, it follows that the degree of
responsibility of $X=1$ for $\ophi$ is at least $1/(n+1)$.  (It is not
hard to show that it is in fact exactly $1/(n+1)$.)  

The argument for showing that checking  if
the degree of responsibility of $X=0$ for $\neg \phi$ is at least
$1/k$ is NP-complete is essentially identical; we leave details to the
reader.  
\eprf}

Again, it follows that the problem of computing the degree of
 responsibility of $X=x$ for $\phi$ in $(M,\vec{u})$ is in \fpnp\ (a little
more effort in the spirit of \cite[Theorem 4.3]{ChocklerH03} shows that it is
\fpnp-complete),
while the problem of computing the degree of blame
of $X=x$ for $\phi$ relative to
an epistemic state
$(\K,\Pr)$ is in
 $\mbox{FP}^{{\rm NP}[n]}$.
\shortv{We can do even better in conjunctive models.}  

\commentout{
  {\bf For \postmin plans (conjunctive formulas) everything is polynomial?}
[[JOE: Yes, that's essentially true.  We should talk about
    \emph{monotone conjunctive models}.  The
    theorem should say that in a 
    monotone conjunctive model, there is a
    unique cause of $\neg \phi$ in $(M,\vec{u})$ \emph{when we
      restrict the conjuncts of the cause to an appropriate set X --
      below)} and that all the   causes of $\phi$ in $(M,\vec{u})$ are
      single conjuncts 
      (``but-for'' causes, in the language of \cite{Hal47}.)
      The ``appropriate set'' has to involve variables that are
      independent (none is an ancestor of another) and form a cut set
      -- every path from an exogenous variable to the formula that is
      caused goes through one of these variables. I think that these
      conditions hold for our application, if we take the appropriate
      to be the intention variables (which are the ones that we're interested
      in), but we should make sure
      that's true.  The fact
    that there is a unique cause means that it follows immediately
    that checking if $X=x$ is part of a cause can be done in
    polynomial time, and computing the (exact) degree of
    responsibility of $X=x$ can also be done in polynomial time---that
    should be a corollary.  But again, for these results, we have to
    restrict to causes in the cutset, which I think we want to do in
    any case.  Let me know if that's clear.]]
}
\fullv{We can do even better in conjunctive models.}
\thm\label{thm:conjunctive} If $M$ is a conjunctive causal model, $\phi$ is a
conjunctive formula, and $(\K,\Pr)$ is an epistemic state where all
the causal models in $\K$ are conjunctive, then the degree of
responsibility of $\vec{X}=\vec{1}$ for $\phi$ (resp., $\vec{X}=\vec{0}$ for
$\neg \phi$) in $(M,\vec{u})$ can be computed in polynomial time, as can 
the degree of blame of 
$\vec{X}=\vec{1}$ for $\phi$ (resp., $\vec{X} = \vec{0}$ for $\neg \phi$) relative 
to $(\K,\Pr)$.
\ethm
\prf 
 It is easy to check that $\vec{X} = \vec{1}$ is a cause of the
conjunctive formula $\phi$ in $(M,\vec{u})$, where $M$ is a
conjunctive causal model, iff $\vec{X}$ is a singleton and 
$(M,\vec{u}) \sat [X=0]\neg \phi$.  (This means $X=1$ a ``but-for''
cause, in legal language.)  Thus, $X=1$ has degree of responsibility 1
for $\phi$.  It is clearly easy to determine if $X=1$ is a but-for
cause of $\phi$ and find all the causes of $\phi$ in polynomial time
in this case.  It follows that the degree of responsibility
and degree of blame of $\vec{X} = \vec{1}$ can also be computed in polynomial time.

In the case of degree of responsibility of $\vec{X}=\vec{0}$ for $\neg \phi$,
observe that  
for a conjunctive formula $\phi$, there is exactly one cause of $\neg \phi$ 
in $(M,\vec{u})$: the one containing all conjuncts of the form $Y=0$. It is 
easy to check whether $X=0$ is part of that single cause, and if it is, then 
its degree of responsibility is $1/k$, where $k$ is the number of variables 
which have value 0. Similarly, it is easy to compute degree of blame in 
polynomial time.
\eprf

Since the causal models \fullv{that are} determined by team plans are
monotone, the upper bounds of Theorem~\ref{thm:monotone} apply
immediately to team plans
(provided that we fix the maximal number of literals in a precondition);
similarly, Theorem~\ref{thm:conjunctive}
applies to team plans that are postcondition minimal.
\shortv{Next we show that the NP-hardness results also apply to team plans.}
\fullv{The question
remains whether the NP-hardness results in parts (d) and (e) of
Theorem~\ref{thm:monotone} also apply to team plans. It is possible
that the causal models that arise from team plans have additional
structure that makes computing whether $X=1$ is part of a cause of
$\phi$ easier than it is for arbitrary monotone
causal models, and similarly 
for responsibility.  As the following result shows, this is not the
case.}


\commentout{
\thm\label{thm:teamplancomplexity} Determining whether $\intd{a}{t}$
is part of a cause of $\perf{Finish}$ in $(M_\P,\vec{u})$, where $M_\P$ is the
causal model determined by a team plan $\P$, is NP-complete, as is
determining whether the degree of responsibility of $\intd{a}{t}$ for
$\perf{Finish}$ is at least $1/k$.
\ethm

\shortv{The proof of the lower bound uses 
Theorem \ref{thm:monotone}(d).  Given a trivial monotone causal model $M$ 
and formula $\phi$ as in Theorem \ref{thm:monotone}(d),
we construct a model $M_\P$ determined by a team plan $\P$ and show
that $X=1$ is part of a cause of $\phi$ in $(M,u)$ iff in $M_\P$
corresponding variable being $1$ is part of a cause of
$\perf{Finish}$.
The full proof can be found in the supplementary material.}

\fullv{
\prf As we observed, the upper bounds follow from parts (d) and (e) of
Theorem~\ref{thm:monotone}.  For the lower bound, recall that it is
already NP-hard to compute whether $X=1$ is part of a cause of $\phi$
in a trivial monotone causal model, where $\phi$
has the form $\psi \land (\phi' \lor X=1)$, $\phi'$
is a formula in DNF whose variables are contained  in $\{X_1,
  \ldots, X_n$, $Y_1$, $\ldots$, $Y_n\}$, and 
  $\psi$ is  the formula   $(X_1 = 1 \lor Y_1 = 1) \land\ldots \land
  (X_n= 1 \lor Y_n = 1)$.  Given such a model $M$ and formula $\phi$,
  we construct a model $M_\P$ determined by a team plan $\P$ as
    follows.  Suppose that $\phi'$ is the formula $\sigma_1 \lor \ldots
  \lor \sigma_k$, where $\sigma_j$ is a conjunction of formulas of the
  form $X_h = 1$ and $Y_h=1$.  The formula $\phi$ is clearly logically
  equivalent to $\phi'' = (\sigma_1 \land \psi) \lor \ldots \lor (\sigma_k
  \land \psi) \lor (X=1 \land \psi)$.  Let $\psi'$ be the formula that
  results by replacing each disjunct $X_i=1 \lor Y_i = 1$ in
    $\psi$ by $W_i = 1$, and let $\phi^*$ be the formula that results from
  replacing each occurrence of $\psi$ in $\phi''$ by $\psi'$.
    Clearly, $\phi^*$ is monotone.

  We construct a team plan $\P =
  (T,Ag,\prec,\alpha,\emptyset)$ with
  $T = \{Start,$ $Finish,t_X, t_{X_1}, \ldots, t_{X_n}, t_{Y_1}, \ldots,
    t_{Y_n}, t_{W_1}, \ldots$, $t_{W_n}\}$; thus, besides $Start$ and
  $Finish$, there is a task corresponding to each
  variable in $\psi'$.  The only nontrivial  ordering conditions are
  $t_{X_i}, t_{Y_i} \prec t_{W_i}$.  Take $Ag=\{a\}$ and take
  $\alpha$ such that 
  each task in $T \setminus \{Start,Finish\}$ is associated with agent $a$.
  Finally, we define $prec$ and $post$ so that $clob(t) = \emptyset$
  for all actions $t$,   $est(t_{W_i}) =
  \{\{t_{X_i}\},\{t_{Y_i}\}\}$, $est(t_{X_i}) = \emptyset$, and
  $est(t_{Y_1}) = \emptyset$   for $i = 1, \ldots, n$, and
  $est(Finish) = \{E_{\sigma_1}, \ldots, E_{\sigma_k}, \{t_{X},
  t_{W_1}, \ldots, t_{W_n}\}\}$, where $E_{\sigma_j}$ consists of the
  tasks $t_{X_i}$ and $t_{Y_j}$ such that $X_i$ and $Y_j$ appear in
  $\sigma_j$, together with $t_{W_1}, \ldots, t_{W_n}$.  This ensures
  that the equation for $\perf{Finish}$ looks like $\phi^*$, except
  each variable $T \in \{X_1,\ldots, X_n,Y_1, \ldots, Y_n,W_1,\ldots,
  W_n\}$ is replaced by $\intd{a}{t_T}$. 

  Consider the
  causal model $M_{\P}$.  We claim that $X=1$ is part of a cause of
  $\phi^*$ in $(M,u)$, where $u$ sets all endogenous variables to 1,
  iff $\intd{a}{t_X}=1$ is a part of a cause of $\perf{Finish}$ in
  $(M_\P,\vec{u}_\P)$, where $\vec{u}_\P$ is such that $\intd{a}{t} =
  1$ for all tasks $t \in T - \{Start,Finish\}$.
  Suppose that $X=1$ is part of a cause of $\phi^*$ in $(M,\vec{u})$.
  Then there exists some $\vec{V} \subseteq \{X_1, \ldots, X_n, Y_1,
  \ldots, Y_n\}$ such that $\vec{V} = \vec{1} \land X=1$ is a cause of
  $\phi^*$.  It is easy to see that the corresponding conjunction
  ($\land_{V \in\vec{V}}\intd{a}{t_V}=1 \land \intd{a}{t_X}=1$) is a cause
  of $\perf{Finish}$ in $(M_\P,\vec{u}_\P)$, so $\intd{a}{t_X}=1$ is
  part of a cause of $\perf{Finish}$. 

  Conversely, suppose that $\intd{a}{t_X}=1$ is part of a cause of
  $\perf{Finish}$ in $(M_\P,\vec{u}_\P)$.  Thus, there exists a set
  $\vec{V}$ such that $\vec{V} = \vec{1} \land \intd{a}{t_X}=1$ is a cause of
  $\perf{Finish}$ in $(M_\P,\vec{u}_\P)$.  Note that
  $\intd{a}{t_{W_i}} \notin 
  \vec{V}$ for $i = 1, \ldots, n$.  For it is easy to see that 
  $(M_\P,\vec{u}_\P) \sat [\intd{a}{t_{W_i}} \gets 0]\neg \perf{Finish}$, so AC3
  would be violated if $\intd{a}{t_{W_i}} \in \vec{V}$. The same holds true
  if $\en{t_{W_i}} \in \vec{V}$ or if $\perf{t_{W_i}} \in \vec{V}$.
  Next note that if $\en{t_T}     \in \vec{V}$ 
    then it can be replaced by $\intd{a}{t_T}$,
    for $T \in \{X_1,\ldots, X_n, Y_1, \ldots, Y_n\}$, and similarly
    for $\perf{t_T}$.  That is, if $\vec{V}'$ is the set obtained
    after doing this replacement, then $\vec{V} \land X=1$ is a cause
    of $\perf{Finish}$ iff $\vec{V}' \land X=1$ is a cause
    of $\perf{Finish}$.  The upshot of this discussion is that,
    without loss of generality, we can take $\vec{V}$ to be a subset of
    $\{\intd{a}{T_T}: T \in \{X_1,\ldots, X_n, Y_1, \ldots, Y_n\}\}$.
    It now easily follows that if $\vec{V}^*$ is the corresponding
    subset of $\{X_1,\ldots, X_n, Y_1, \ldots, Y_n\}$, then $\vec{V}^*
    = \vec{1} 
\land X=1$ is a cause of $\phi^*$ in $(M,u)$.  This completes the proof.
The argument in the case of responsibility uses
Theorem~\ref{thm:monotone}(e) in the same way, and is left to the reader.
    \eprf
}
}

\thm\label{thm:teamplancomplexity} 
Determining whether $\intd{a}{t}=0$
is part of a cause of $\neg \perf{Finish}$ in $(M_\P,\vec{u})$, where 
$M_\P$ is the causal model determined by a team plan $\P$, is NP-complete, 
as is determining whether the degree of responsibility of agent $a$ for
$\neg \perf{Finish}$ is at least $m/k$.
\ethm

\shortv{The proof of the lower bound uses 
Theorem \ref{thm:monotone}(d).  Given a trivial monotone causal model $M$ 
and formula $\phi$ as in Theorem \ref{thm:monotone}(d),
we construct a model $M_\P$ determined by a team plan $\P$ and show
that $X=0$ is part of a cause of $\neg \phi$ in $(M,u)$ iff in $M_\P$
corresponding variable being $0$ is part of a cause of
$\neg \perf{Finish}$.
The full proof can be found in the supplementary material.}

\fullv{
\prf As we observed, the upper bound for determining whether $\intd{a}{t}=0$
is part of a cause follows from part (d) of Theorem~\ref{thm:monotone}.  
For the lower bound, recall that it is
already NP-hard to compute whether $X=0$ is part of a cause of $\neg \phi$
in a trivial monotone causal model, where $\phi$
has the form $\psi \land (\phi' \lor X=1)$, $\phi'$
is a formula in DNF whose variables are contained  in $\{X_1,
  \ldots, X_n, Y_1, \ldots$, $Y_n\}$, and 
  $\psi$ is  the formula   $(X_1 = 1 \lor Y_1 = 1) \land\ldots \land
  (X_n= 1 \lor Y_n = 1)$.  Given such a model $M$ and formula $\phi$,
  we construct a model $M_\P$ determined by a team plan $\P$ as
    follows.  Suppose that $\phi'$ is the formula $\sigma_1 \lor \ldots
  \lor \sigma_k$, where $\sigma_j$ is a conjunction of formulas of the
  form $X_h = 1$ and $Y_h=1$.  The formula $\phi$ is clearly logically
  equivalent to $\phi'' = (\sigma_1 \land \psi) \lor \ldots \lor (\sigma_k
  \land \psi) \lor (X=1 \land \psi)$.  Let $\psi'$ be the formula that
  results by replacing each disjunct $X_i=1 \lor Y_i = 1$ in
    $\psi$ by $W_i = 1$, and let $\phi^*$ be the formula that results from
  replacing each occurrence of $\psi$ in $\phi''$ by $\psi'$.
    Clearly, $\phi^*$ is monotone. 

        We construct a team plan $\P = (T,Ag,\prec,\alpha)$ with 
  $T = \{Start,$ $Finish,t_X, t_{X_1}, \ldots, t_{X_n}, t_{Y_1}, \ldots,
                t_{Y_n}, t_{W_1}, \ldots$, $t_{W_n}\}$; that is, besides $Start$ and
  $Finish$, there is a task corresponding to each
  variable in $\psi'$.  The only nontrivial  ordering conditions are
  $t_{X_i}, t_{Y_i} \prec t_{W_i}$.  
Take $Ag=\{a_t: t \in T \setminus \{Start,Finish\}\}$ and take
  $\alpha$ such that 
   each task $t$ in $T \setminus \{Start,Finish\}$ is associated with agent $a_t$.
  Finally, we define $prec$ and $post$ so that $clob(t) = \emptyset$
  for all actions $t$,   $est(t_{W_i}) =
  \{\{t_{X_i}\},\{t_{Y_i}\}\}$, $est(t_{X_i}) = \emptyset$, and
  $est(t_{Y_1}) = \emptyset$   for $i = 1, \ldots, n$, and
  $est(Finish) = \{E_{\sigma_1}, \ldots, E_{\sigma_k}$, $\{t_{X},
  t_{W_1}$, $\ldots, t_{W_n}\}\}$, where $E_{\sigma_j}$ consists of the
  tasks $t_{X_i}$ and $t_{Y_j}$ such that $X_i$ and $Y_j$ appear in
  $\sigma_j$, together with $t_{W_1}, \ldots, t_{W_n}$.  This ensures
  that the equation for $\perf{Finish}$ looks like $\phi^*$, except
  each variable $Z \in \{X_1,\ldots, X_n,Y_1, \ldots, Y_n,W_1,\ldots,
  W_n\}$ is replaced by $\intd{a_{t_Z}}{t_Z}$. 

  Consider the
  causal model $M_{\P}$.  We claim that $X=0$ is part of a cause of
  $\neg \phi^*$ in $(M,u)$, where $u$ sets all endogenous variables to 0,
  iff $\intd{a_{t_X}}{t_X}=0$ is a part of a cause of $\neg \perf{Finish}$ in
  $(M_\P,\vec{u}_\P)$, where $\vec{u}_\P$ is such that $\intd{a_t}{t} =
  0$ for all tasks $t \in T \setminus \{Start,Finish\}$.
  Suppose that $X=0$ is part of a cause of $\neg \phi^*$ in $(M,\vec{u})$.
  Then there exists some $\vec{V} \subseteq \{X_1, \ldots, X_n, Y_1,
  \ldots, Y_n\}$ such that $\vec{V} = \vec{0} \land X=0$ is a cause of
    $\neg \phi^*$.  The corresponding conjunction
  ($\land_{V \in\vec{V}}\intd{a_{t_V}}{t_V}=0 \land \intd{a_{t_X}}{t_X}=0$) is a cause
  of $\neg \perf{Finish}$ in $(M_\P,\vec{u}_\P)$, so $\intd{a_{t_X}}{t_X}=0$ is
  part of a cause of $\neg \perf{Finish}$. 

  Conversely, suppose that $\intd{a_{t_X}}{t_X}=0$ is part of a cause of
  $\neg \perf{Finish}$ in $(M_\P,\vec{u}_\P)$.  Thus, there exists a set
  $\vec{V}$ such that $\vec{V} = \vec{0} \land \intd{a_{t_X}}{t_X}=0$ is a cause of
  $\neg \perf{Finish}$ in $(M_\P,\vec{u}_\P)$.  Note that
  $\intd{a_{t_{W_i}}}{t_{W_i}} \notin 
  \vec{V}$ for $i = 1, \ldots, n$.  For it is easy to see that 
  $(M_\P,\vec{u}_\P) \sat [\intd{a_{t_{W_i}}}{t_{W_i}} \gets 1]\perf{Finish}$, so AC3
  would be violated if $\intd{a_{t_{W_i}}}{t_{W_i}} \in \vec{V}$. The same holds true
  if $\en{t_{W_i}} \in \vec{V}$ or if $\perf{t_{W_i}} \in \vec{V}$.
  Next note that if $\en{t_Z}     \in \vec{V}$ 
    then it can be replaced by $\intd{a_{t_Z}}{t_Z}$,
    for $Z \in \{X_1,\ldots, X_n, Y_1, \ldots, Y_n\}$, and similarly
    for $\perf{t_Z}$.  That is, if $\vec{V}'$ is the set obtained
    after doing this replacement, then $\vec{V} \land X=0$ is a cause
    of $\neg \perf{Finish}$ iff $\vec{V}' \land X=0$ is a cause
    of $\neg \perf{Finish}$.  The upshot of this discussion is that,
    without loss of generality, we can take $\vec{V}$ to be a subset of
    $\{\intd{a_{t_Z}}{t_Z}: Z \in \{X_1,\ldots, X_n, Y_1, \ldots, Y_n\}\}$.
    It now easily follows that if $\vec{V}^*$ is the corresponding
    subset of $\{X_1,\ldots, X_n, Y_1, \ldots, Y_n\}$, then $\vec{V}^*
    = \vec{0} 
    \land X=0$ is a cause of $\neg \phi^*$ in $(M,u)$.  This completes 
    the proof for part of a cause.

The argument for the degree of responsibility is similar to
Theorem~\ref{thm:monotone}(e). For the upper bound, we guess a cause
where the proportion of  
$a$-controlled variables with value 0 is greater or equal to
$m/k$. Then we can check in polynomial time that it is indeed a cause
of $\neg \perf{Finish}$. The lower bound follows from the previous
argument (for the special case when 
$m=1$ and the degree of responsibility of an agent $a_t$ is the same as
the degree of responsibility of $\intd{a_t}{t}$), as in 
Theorem~\ref{thm:monotone}(e).
    \eprf
} 


\section{Conclusions}\label{sec:conclusions}
We have shown how the definitions of causality, responsibility and blame from 
\cite{Hal47} can be used to give useful insights in the context of team plans. 
We also showed that the resulting problems are tractable:
causality
for team plans can be computed in polynomial time, while the problem of
determining the degree of responsibility and blame is NP-complete; for \postmin plans, the degree of
responsibility and blame can be computed in polynomial time.
We can extend our model with external events (or actions by an environment
agent) without increase in complexity. We chose not to consider events here,
as we are concerned only with allocating responsibility and blame to
agents (rather than to the environment). In future work, we would like to
consider a richer setting, where agents may be able to perform actions that
decrease the probability of plan failure due to external events.

The epistemic perspective of the paper is that of an outside observer
rather than the agents. In future work we plan to model agents
reasoning about the progress of plan execution, which would involve
their beliefs about what is happening and who is to blame for the
failure of the plan.  

  
\bibliographystyle{abbrv}

\end{document}